\documentclass[10pt,twocolumn,letterpaper]{article}

\usepackage{iccv}
\usepackage{times}
\usepackage{epsfig}
\usepackage{graphicx}
\usepackage{amsmath}
\usepackage{amssymb}

\usepackage{booktabs}
\usepackage{multirow}
\usepackage{graphicx}
\usepackage{amsmath}
\usepackage{bbm}
\usepackage{booktabs}
\usepackage{placeins}
\usepackage[dvipsnames]{xcolor}
\usepackage{algorithm}
\usepackage[noend]{algpseudocode}
\usepackage{wrapfig}
\usepackage[pagebackref=true,breaklinks=true,letterpaper=true,colorlinks,bookmarks=false]{hyperref}

\usepackage[capitalize]{cleveref}
\crefname{section}{Sec.}{Secs.}
\Crefname{section}{Section}{Sections}
\Crefname{table}{Table}{Tables}
\crefname{table}{Tab.}{Tabs.}

\def\eg{\emph{e.g}\onedot} 
\def\ie{\emph{i.e}\onedot} 
 
\def\etc{\emph{etc}\onedot} \def\vs{\emph{vs}\onedot}
 
\def\etal{\emph{et al}\onedot}

\newcommand{\cutsectionup}{\vspace*{-8pt}} 
\newcommand{\cutsectiondown}{\vspace*{-2pt}}
\newcommand{\cutsubsectionup}{\vspace*{-3pt}}

\newcommand{\cutparagraphup}{\vspace*{-10pt}}

\newcommand{\cutcaptionup}{\vspace*{-6pt}}
\newcommand{\cutcaptiondown}{\vspace*{-2pt}}

\newcommand{\eccvname}{MI-UP}
\newcommand*\samethanks[1][\value{footnote}]{\footnotemark[#1]}

\iccvfinalcopy 

\begin{document}

\title{Unsupervised 3D Perception with 2D Vision-Language Distillation \\ for Autonomous Driving}

\author{Mahyar Najibi\thanks{Equal contribution} \quad
Jingwei Ji\samethanks \quad
Yin Zhou\thanks{Corresponding author} \quad Charles R. Qi \quad Xinchen Yan
\\ Scott Ettinger \quad Dragomir Anguelov
\\ Waymo LLC}

\maketitle
\ificcvfinal\thispagestyle{empty}\fi

\begin{abstract}
Closed-set 3D perception models trained on only a pre-defined set of object categories can be inadequate for safety critical applications such as autonomous driving where new object types can be encountered after deployment. In this paper, we present a multi-modal auto labeling pipeline capable of generating amodal 3D bounding boxes and tracklets for training models on open-set categories without 3D human labels. Our pipeline exploits motion cues inherent in point cloud sequences in combination with the freely available 2D image-text pairs to identify and track all traffic participants. Compared to the recent studies in this domain, which can only provide class-agnostic auto labels limited to moving objects, our method can handle both static and moving objects in the unsupervised manner and is able to output open-vocabulary semantic labels thanks to the proposed vision-language knowledge distillation. Experiments on the Waymo Open Dataset show that our approach outperforms the prior work by significant margins on various unsupervised 3D perception tasks.
\end{abstract}

\cutsectionup
\section{Introduction} \label{sec:intro}
\cutsectiondown
In autonomous driving, most existing 3D detection
models~\cite{zhou2018voxelnet,REF:pointpillars_cvpr2018,shi2019pointrcnn} have been developed with the prior assumption that all possible categories of interest should be known and annotated during training.
While significant progress has been made in this supervised closed-set setting, these methods still struggle to fully address the safety concerns that arise in high-stakes applications.
Specifically, 
in the dynamic real-world environment,
it is unacceptable for autonomous vehicles to fail to handle a category that is not present in the training data.
To address this safety concern, a recent development by Najibi~\etal\cite{najibi2022motion} proposed an unsupervised auto labeling pipeline that uses motion cues from point cloud sequences to localize 3D objects. 
However, by design, this method does not localize static objects which constitute a significant portion of traffic participants. Moreover, it only models the problem in a class-agnostic way and fails to provide semantic labels for scene understanding. This is suboptimal as semantic information is essential for downstream tasks such as motion planning, where category-specific safety protocols are deliberately added to navigate through various traffic participants.

\begin{figure}
\includegraphics[width=\linewidth]{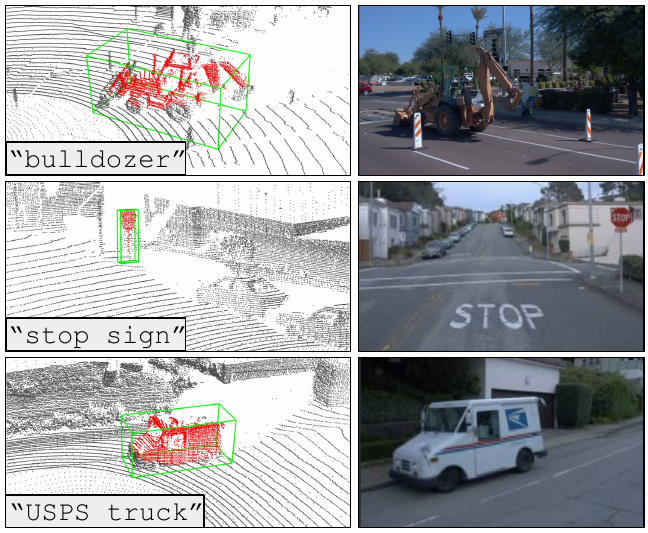}
\cutcaptionup
\cutcaptionup
\caption{An illustration of three interesting urban scene examples of open-vocabulary perception. 
Left: our method can faithfully detect objects based on user-provided text queries during inference, without the need for 3D human supervision. Red points are points matched with the text queries. Right: camera images for readers' reference. Note that the inference process solely relies on LiDAR points and does not require camera images.}
\cutcaptiondown
\label{fig:teaser}
\end{figure}

Recently, models trained with large-scale image-text datasets have demonstrated robust flexibility and generalization capabilities for open-vocabulary image-based classification \cite{radford2021learning,jia2021scaling,ming2022delving}, detection \cite{kamath2021mdetr,gu2021open,li2022grounded,zhong2022regionclip} and semantic segmentation \cite{li2022language,ghiasi2022scaling} tasks. 
Yet, open-vocabulary recognition in the 3D domain~\cite{fan2022minedojo,huang2022inner,shafiullah2022clip} is in its early stages. In the context of autonomous driving it is even more underexplored.
In this work, we fill this gap by leveraging a pre-trained vision-language model to realize open-vocabulary 3D perception in the wild.

We propose a novel paradigm of \underline{U}nsupervised 3D \underline{P}erception with 2D \underline{V}ision-\underline{L}anguage distillation (UP-VL). 
Specifically, by incorporating a pre-trained vision-language model, UP-VL can generate auto labels with substantially higher quality for objects in arbitrary motion states, compared to the latest work by Najibi~\etal\cite{najibi2022motion}.

With our auto labels,
we propose to co-train a 3D object detector with a knowledge distillation task, which can achieve two goals simultaneously, \ie improving detection quality and transferring semantic features from 2D image pixels to 3D LiDAR points.
The perception model therefore is capable of detecting all traffic participants
and thanks to the distilled open-vocabulary features, we can flexibly query the detector's output embedding with text prompts, for preserving specific types of objects at inference time (see Figure \ref{fig:teaser} for some examples). 

We summarize the contributions of UP-VL as follows:

\begin{itemize}
    \item UP-VL achieves state-of-the-art performance on unsupervised 3D perception (detection and tracking) of moving objects for autonomous driving.
    \item UP-VL introduces semantic-aware unsupervised detection for objects in any motion state, a first in the field of autonomous driving. This breakthrough eliminates the information bottleneck that has plagued previous work~\cite{najibi2022motion}, where class-agnostic auto labels were used, covering only moving objects with a speed above a predetermined threshold.
    \item UP-VL enables 3D open-vocabulary detection of novel objects in the wild, with queries specified by users at inference time, therefore removing the need to re-collect data or re-train models.
\end{itemize}

\cutsectionup
\section{Related works} 
\cutsectiondown
\label{sec:related_works}

\paragraph{Vision-language training.}
Contrastive vision language training on billions of image-text training pairs resulted in impressive improvements in the tasks of open-set and zero-shot image classification and language related applications \cite{radford2021learning,jia2021scaling,zhai2022lit}. More recently, open-set object localization in 2D images has been shown to benefit from such abundant image-text data as well. Specifically, \cite{kamath2021mdetr,gu2021open,li2022grounded,zhong2022regionclip,zhou2022detecting,minderer2022simple} used image-text training to improve the open-set capability of 2D object detectors and \cite{li2022language,ghiasi2022scaling} explored the use of large-scale scene-level vision-language data for the task of open-set 2D semantic segmentation.
Recent research~\cite{lu2022open, kobayashi2022decomposing,tschernezki2022neural,ha2022semantic, jain2022bottom} has begun to explore the application of 2D vision-language pre-training in 3D perception tasks. 
However, these studies focused on static indoor scenario where the scene is small-scale and the RGB-D data is captured in high-resolution.
Here we design a multi-modal pipeline that leverages vision-language pre-training for unsupervised open-set 3D perception in complex, sparse, and occlusion-rich environments for autonomous driving.

\cutparagraphup
\paragraph{Unsupervised 3D object detection.}
Unsupervised 3D object detection from LiDAR data is largely under-explored~\cite{dewan2016motion,wong2020identifying,tian2021unsupervised,liu2021opening,najibi2022motion}.
Dewan~\etal~\cite{dewan2016motion} proposed a model-free method to detect and track the visible part of objects, by using the motion cues from LiDAR sequences. However, this approach is incapable of generating amodal bounding boxes which is essential for autonomous driving.
Cen~\etal~\cite{cen2021open} relied on a supervised detector to produce proposals of unknown categories. However, this approach requires full supervision to train the base detector and has limited generalization capability to only semantically similar categories.
Wong~\etal~\cite{wong2020identifying} identified unknown instances via supervised segmentation and clustering, which by design cannot generate amodal boxes from partial observations.
Most recently, Najibi~\etal~\cite{najibi2022motion} developed an unsupervised auto meta labeling pipeline to generate pseudo labels for moving objects, which can be used to train real-time 3D detection models. This approach fails to provide semantics to detection boxes and ignores static objects, which limits its practical utility. 
Compared to all previous efforts, we realize open-vocabulary unsupervised 3D detection for both static and moving objects, by leveraging vision-language pre-training, and benchmark our system on the realistic and challenging scenario of autonomous driving.
While utilizing 2D vision-language models that may have been pre-trained with human annotations, we avoid the need for any additional 3D labels within our paradigm, thereby creating a pragmatically unsupervised setting.

\cutparagraphup
\paragraph{LiDAR 3D object detection.}
Most previous works focused on developing performant model architectures in the fully supervised setting, without considering the generalization capability to long-tail cases and unknown object types that are prevalent in the dynamic real world. These methods can be categorized into point based~\cite{shi2019pointrcnn,qi2019deep,REF:Yang_3DSSD_2020_CVPR,REF:Point-GNN_CVPR2020,misra2021end,li2021lidar}, voxelization based~\cite{REF:Vote3Deep_ICRA2017,REF:VotingforVoting_RSS2015,song2016deep,najibi2020dops,yang2018pixor,REF:simony2018complex,zhou2018voxelnet,REF:pointpillars_cvpr2018,REF:PillarNet_ECCV2020,REF:HVNet2020,zheng2021se,chen2022focal,liu2022bevfusion}, perspective projection based~\cite{meyer2019lasernet,REF:bewley2020range,fan2021rangedet}, and feature fusion ~\cite{sun2021rsn,REF:FastPointRCNN_Jiaya_ICCV2019,zhou2020end,REF:SA_SSD_He_2020_CVPR,shi2020pv}. Recent research also explore transferring knowledge from image for 3D point cloud understanding~\cite{sautier2022image,liu2021learning,janda2022self,chen2023clip2scene}. Our method is compatible with any 3D detector, extending it to handle the open-set settings.

\begin{figure*}
    \centering
    \includegraphics[width=\linewidth]{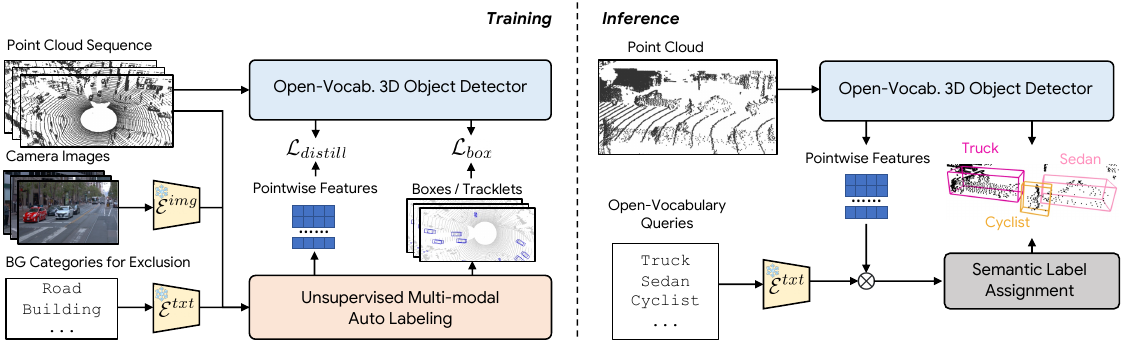}
    \caption{Overview of the proposed UP-VL framework. During training (left), our method taps into multi-modal inputs (LiDAR, camera, text) and produces high-quality auto supervisions, via Unsupervised Multi-modal Auto Labeling, including 3D point-level features, 3D object-level bounding boxes and tracklets. Our auto labels are then used to supervise a class-agnostic open-vocabulary 3D detector. Besides, our 3D detector distills the features extracted from a pre-trained 2D vision-language model. At inference time (right), our trained 3D detector produces class-agnostic boxes and per-point features in the embedding space of the pre-trained vision-language model. We then use the text encoder to map queries to the embedding space and compute the per-point similarity scores between the predicted feature and the text embeddings ($\otimes$ refers to cosine similarity). These per-point scores are then aggregated to assign semantic labels to boxes. 
    }
    \label{fig:top_level_overview}
\end{figure*}

\section{Method}
\cutsectiondown
\label{sec:method}

We present UP-VL, a new approach for unsupervised open-vocabulary 3D detection and tracking of traffic participants. 
UP-VL advances the previous state-of-the-art~\cite{najibi2022motion} which was limited to \emph{class-agnostic} detection of \emph{moving-only} objects in two main directions:
1) It enables \emph{class-aware} open-set 3D detection by incorporating open-vocabulary text queries at inference time, and
2) It is able to detect objects in \emph{all motion states} as opposed to moving-only objects in the previous study.
To achieve these goals, we deploy a multi-modal approach and combine intrinsic motion cues~\cite{najibi2022motion} available from the LiDAR sequences with the semantics captured by a vision-language model~\cite{ghiasi2022scaling} trained on generic image-text pairs from the Internet.
An overview of our approach is shown in Figure~\ref{fig:top_level_overview}. As illustrated on the left, our training pipeline involves two main stages. First, our auto labeling method uses these motion and semantic cues to automatically label the raw sensor data, yielding class-agnostic 3D bounding boxes and tracklets as well as point-wise semantic features. Then, in the second stage, we use these auto labels to train open-vocabulary 3D perception models. The right side of the figure illustrates our inference pipeline where given raw LiDAR point clouds, our detector is able to perform open-vocabulary 3D detection given a set of text queries.

\subsection{Background}
\cutsectiondown
\label{method:background}
The key challenges in unsupervised 3D perception are twofold: 1) generating high-quality 3D amodal bounding boxes and consistent tracklets for all open-set traffic participants, and 2) inferring per-object semantics. 
Najibi~\etal\cite{najibi2022motion} developed an auto labeling technique to address the first challenge partially. Their approach focuses on moving objects only.
Specifically, their method takes LiDAR sequences as input, and removes ground points.
It then breaks down the scene into individual connected components (\ie point clusters).
Next, it calculates local flow between pairs of point clusters from two adjacent frames and retains only clusters with speed above a predefined threshold.
It then tracks each cluster across frames and aggregates points to obtain a more comprehensive view of the object, which enables the derivation of a faithful 3D amodal bounding box.
Finally, the resulting 3D amodal boxes and tracklets can serve as auto labels for training 3D perception models.

While the previous work~\cite{najibi2022motion} has shown promising results, it suffers from significant limitations: 1) it can only deal with moving objects; and 2) it is unable to output semantics. These limitations hinder its practical utility for safety-critical applications such as autonomous driving.

\subsection{Unsupervised Multi-modal Auto Labeling}
\label{method:AML}

In contrast to the traditional way of training a detection model by presenting box geometries and closed-set semantics, our unsupervised multi-modal auto labeling approach produces box geometries and point-wise semantic feature embeddings, where the former teaches the detector to localize all traffic participants and the latter informs the model to preserve certain types of objects based on the inference-time text queries.

\begin{figure}[!t]
\centering
\includegraphics[width=0.95\linewidth]{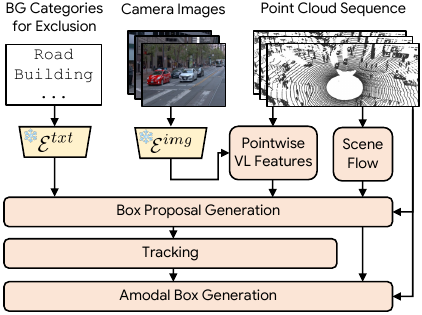}

\caption{Overview of our unsupervised multi-modal auto labeling approach. This pipeline first extracts vision-language and motion features from multiple modalities, then proposes, tracks and completes bounding boxes of objects. The resulting pointwise VL features, 3D bounding boxes and tracklets will serve as automatic supervisions to train the perception model. 
}
\vspace{-1em}
\label{fig:auto_labeling}
\end{figure}

Figure~\ref{fig:auto_labeling} shows an overview of the auto labeling pipeline and Algorithm~\ref{alg:auto_labeling} presents its details. Specifically, our system leverages multiple modalities as input, namely camera images, LiDAR point sequences, and natural language. It also employs a pre-trained vision-language model~\cite{ghiasi2022scaling} to extract feature embeddings from images and texts, which naturally complements the 3D depth information and motion cues with rich semantics, compared to~\cite{najibi2022motion}. We begin by detailing the feature extraction process. We then describe how we utilize the extracted vision-language information in combination with the inherent motion cues from LiDAR sequences to generate auto labels in an unsupervised manner.  

\vspace{3mm}\noindent\textbf{Feature Extraction}\vspace{1mm}

As the first step to our approach, we start by extracting open-vocabulary features from all available cameras and then transfer these 2D features to 3D LiDAR points using known sensor calibrations. Specifically, at each time $t$, we have a set of images $\{\mathbf{I}_t^k \in \mathbb{R}^{H_k\times W_k \times 3}\}_t$ captured by $K$ cameras, where $H_k$ and $W_k$ are image dimensions of the camera $k$. We also have a collection of point cloud, $\{\mathbf{P}_t \in \mathbb{R}^{N_t \times 3}\}$, captured over time using LiDAR sensors. Here, $N_t$ denotes the number of points at time $t$. We use a pre-trained open-vocabulary 2D image encoder $\mathcal{E}^{img}$ to extract the pixel-wise visual features for each image, denoted as $\{\mathbf{V}_t^k \in \mathbb{R}^{H_k \times W_k \times D}\}$, where $D$ represents the feature dimension. 
Next, we build the mapping between 3D LiDAR points and their corresponding image pixels using the camera and LiDAR calibration information. Once this mapping is created, we can associate each 3D point with its corresponding image feature vector.
As a result, we obtain vision-language features for all the 3D points as $\mathbf{F}^{vl}_t \in \mathbb{R}^{N_t \times D}$, where $N_t$ is the number of points at time $t$.

Additionally, we leverage motion signals as another crucial representation that can substantially aid in deducing the concept of objectness for moving instances in the open-set environment.
Specifically, we employ the NSFP++ algorithm~\cite{najibi2022motion} to compute the scene flow $\mathbf{F}^{sf}_t \in \mathbb{R}^{N_t \times 3}$ of points at each time $t$, which is a set of flow vectors corresponding to each point in $\mathbf{P}_t$.

\begin{algorithm}[!ht]
\resizebox{\linewidth}{!}{%
\begin{minipage}{\linewidth}

\caption{\small{Unsupervised multi-modal auto labeling.\label{alg:auto_labeling}}}
\textbf{Input:} A sequence of images across $T$ frames for each of the $K$ cameras $\{\mathbf{I}^k_t\}$; a sequence of LiDAR point locations $\{\mathbf{P}_t\}$. \\
\textbf{Requires:} Cosine similarity threshold for background categories $\epsilon^{bg}$; minimum scene flow magnitude $\epsilon^{sf}$; maximum ratio of background points within a box $r^{bg}$; a set of a prior background categories $\mathbf{C}^{bg}$; a pre-trained open-vocabulary model with image encoder $\mathcal{E}^{img}$ and text encoder $\mathcal{E}^{txt}$.  \\
\textbf{Output:} Amodal 3D bounding boxes $\{\mathbf{B}_t\}$ and their track IDs $\{\mathbf{T}_t\}$; point-wise open-vocabulary features $\{\mathbf{F}^{vl}_t\}$. \\
\textbf{Function:}
\begin{algorithmic}[1]

\For{$t=1$ to $T$}
    \State $\{\mathbf{V}_{t}^k\} \gets \mathcal{E}^{img}(\{\mathbf{I}_{t}^k\})$  \Comment 2D VL features
    \State $\mathbf{F}^{vl}_t \gets \text{Unprojection}(\{\mathbf{V}_t^k\}, \mathbf{P}_t)$  \Comment 3D VL features
    \If{$t \neq T$}
        \State $\mathbf{F}^{sf}_t \gets \text{NSFP++}(\mathbf{P}_t, \mathbf{P}_{t+1})$  \Comment Scene flow
    \Else
        \State $\mathbf{F}^{sf}_t \gets -\text{NSFP++}(\mathbf{P}_t, \mathbf{P}_{t-1})$
    \EndIf
    \For{$i=1$ to $N_t$}
        \State $(\mathbf{M}^{sf}_t)_i \gets \mathbbm{1}(\|(\mathbf{F}^{sf}_t)_i\| \geq \epsilon^{sf})$
        \State $(\mathbf{M}^{bg}_t)_i \gets \mathbbm{1}(\max\limits_{c \in \mathbf{C}^{\text{bg}}}\frac{(\mathbf{F}^{vl}_t)_i \cdot \mathcal{E}^{txt}(c)}{\|(\mathbf{F}^{vl}_t)_i\| \|\mathcal{E}^{txt}(c)\|} \geq \epsilon^{bg})$

    \EndFor
    \State $\widetilde{\mathbf{P}}_t, \widetilde{\mathbf{F}}^{sf}_t \gets \mathbf{P}_t[\mathbf{M}^{sf}_t], \mathbf{F}^{sf}_t[\mathbf{M}^{sf}_t]$
    \State $\mathbf{B}^{vis}_t \gets \text{InitialBoxProposal}(\widetilde{\mathbf{P}}_t, \widetilde{\mathbf{F}}^{sf}_t, \mathbf{M}^{bg}_t; r^{bg})$
\EndFor
\State $\{\mathbf{T}_t\} \gets \text{Tracking}(\{\mathbf{B}^{vis}_t\})$
\State $\{\mathbf{B}_t\} \gets \text{AmodalBoxGeneration}(\{\mathbf{B}^{vis}_t\}, \{\mathbf{T}_t\}, \{\mathbf{P}_t)\}$
\\
\Return $\{\mathbf{B}_t\}$, $\{\mathbf{T}_t\}$, $\{\mathbf{F}^{vl}_t\}$
\end{algorithmic}
\end{minipage}%
}
\end{algorithm}

\vspace{2mm}
\noindent\textbf{Bounding Box Proposal Generation}
\vspace{1mm}

At each time step, we generate initial bounding box proposals $\{\mathbf{B}^{vis}_t \in \mathbb{R}^{M_t \times 7}\}$ by clustering the points, where $M_t$ is the number of boxes at time $t$, and each box is parameterized as (center $x$, center $y$, center $z$, length, width, height, heading). Note that $vis$ indicates that each box only covers the visible portion of an object.
To cluster each point, we leverage a set of features which includes the point locations $\mathbf{P_t}$, scene flow $\mathbf{F}^{sf}_t$, and the vision-language features $\mathbf{F}^{vl}_t$.

We design our pipeline to flexibly generate auto labels for objects in desired motion states. Given scene flow $\mathbf{F}^{sf}_t$, we introduce a velocity threshold $\epsilon^{sf}$ to select points whose speed is greater than or equal to the threshold (\eg, 1.0 m/s). To capture objects in all motion states, we set $\epsilon^{sf} = 0$.

One major challenge of auto labeling objects in all motion states is how to automatically distinguish traffic participants (\eg, vehicles, pedestrians, \etc) from irrelevant scene elements (\eg, street, fence, \etc). 
We propose to leverage an a priori list of \textit{background} object categories to exclude irrelevant scene elements from labeling.
Specifically, we use the text encoder, $\mathcal{E}^{txt}$, from the pre-trained 2D vision-language model~\cite{ghiasi2022scaling}, to encode each background category name $c$ into its feature embedding $\mathcal{E}^{txt}(c) \in \mathbb{R}^D$. We further define a per-point binary background mask, denoted as $\mathbf{M}^{bg}_t \in \{0, 1\}^{N_t}$, that takes on a value of 1 if a point is assigned to one of the a priori background categories, or 0 otherwise. See Algorithm \ref{alg:auto_labeling} for the definition of $\mathbf{M}^{bg}_t$, where $(\cdot)_i$ denotes the $i$-th row of a matrix and $\mathbbm{1}(\cdot)$ represents the indicator function. We use this background mask to mark scene elements which are not of interest.

We then proceed to cluster the point cloud into neighboring regions using a spatio-temporal clustering algorithm, modified from~\cite{najibi2022motion}, followed by calculating the tightest bounding box around each cluster.
In addition to clustering points by their locations and motions, we also use $\mathbf{M}^{bg}_t$ to eliminate bounding boxes which are likely to be background. To be precise, we discard any bounding box in which the ratio of background points exceeds a threshold of $r^{bg}$ (which is set to 99\%). This process results in the initial set of bounding box proposals $\{\mathbf{B}^{vis}_t\}$. 
Note that in this step, the box dimensions are determined based on the \textit{visible} portion of each object, which can be significantly underestimated compared to the human labeled amodal box, due to ubiquitous occlusions and sparsity.

\vspace{2mm}
\noindent\textbf{Amodal Auto Labeling}
\vspace{1mm}

In autonomous driving, perception downstream tasks desire \textit{amodal} boxes that encompass both the visible and occluded parts of the objects.
To transform our visible-only proposals to amodal auto labels, we follow~\cite{najibi2022motion} by adopting a tracking-by-detection paradigm with Kalman filter state updates to link all proposals over time.
We then perform shape registration for each object track of $\{\mathbf{T}_t\}$ using ICP~\cite{besl1992method}.
Within each track, we leverage the intuition that different viewpoints contain complementary information and temporal aggregation of the registered points from proposals would allow us to obtain a complete shape of the object. Hence, we fit a new box to the aggregated points to yield the amodal box.
Finally, we undo the registration from aggregated points to individual frames and replace the original visible box proposal at each time step with the amodal box, which produces auto labeled 3D boxes and the tracklet.

In practice, background point filtering, point cloud registration and temporal aggregation may contain noise, leading to spurious boxes, \eg, tiny and sizable boxes and overlapping boxes. We apply non-maximum suppression (NMS) to clean the auto label boxes.
This final set of unsupervised amodal auto labels $\{\mathbf{B}_t\}$, their track IDs $\{\mathbf{T}_t\}$, together with the extracted vision-language embeddings $\{\mathbf{F}^{vl}_t\}$, are then used to train open-vocabulary 3D object detection model as described in Sec. \ref{sec:method_VL}.

\cutsubsectionup
\subsection{Open-vocabulary 3D Object Detection}
\label{sec:method_VL}

In this subsection, we describe how the unsupervised auto labels, can be used to train a 3D object detector capable of localizing open-set objects and assigning open-vocabulary semantics to them, all without using any 3D human annotations during training.

\subsubsection{Model Architecture}
\label{sec:det_model_arch}

Our design, as depicted in Figure~\ref{fig:top_level_overview}, is based on decoupling object detection into class-agnostic object localization and semantic label assignment.
For class-agnostic bounding box prediction, we add a branch to a 3D point cloud encoder backbone to generate 3D bounding box center, dimensions, and heading. 
This branch accompanies a binary classification branch which outputs foreground / background class-agnostic per box objectness score. 
To supervise these two branches, we treat our unsupervised auto labels (see Sec.~\ref{method:AML}) as ground-truth and add bounding box regression and classification losses to our learning objective.
We would like to highlight that our pipeline is independent of a specific 3D point-cloud encoder~\cite{ REF:pointpillars_cvpr2018, zhou2020end, SWFormer} and the detection paradigm (either anchor-based or anchor-free detection). 
Here, we adopt an anchor-based PointPillar backbone \cite{REF:pointpillars_cvpr2018} with Huber loss for box residual regression and Focal Loss~\cite{lin2017focal} for objectness classification to have a fair comparison with prior works~\cite{najibi2022motion}. 
Besides predicting 3D bounding boxes, we also perform text query-based open-vocabulary semantic assignment by distilling knowledge from pre-trained 2D vision-language models using an extra branch which is described in the next subsections. 

\subsubsection{Vision-Language Knowledge Distillation}
\label{sec:distillation}

Besides class agnostic bounding box generation, our 3D detector pipeline also distills the semantic knowledge from the per-point vision-language features provided by our auto labeling pipeline (\ie $\{\mathbf{F}^{vl}_t\}$, introduced in the the vision-language feature extraction in Sec.~\ref{method:AML}).
In our method, we directly distill these features, which as will be discussed in the next subsection, unlocks text query-based open-vocabulary category assignment at inference time. More precisely, as shown in the left side of Figure~\ref{fig:top_level_overview}, we add a new linear branch to the model to predict per-point $D$ dimensional features (here $D$ is the dimensionality of the vision-language embedding space). As the input to this branch, we scatter the computed voxelized features in our backbone back into the points and concatenate them with the available per-point input features (\ie 3D point locations and LiDAR intensity and elongation features). We then train the network to predict the feature vector $\mathbf{f}_p^{vl} \in \mathbf{F}^{vl}_t$ for any point $\mathbf{p}$ visible in the camera images and add the following loss to the training objective:

\begin{equation}
    \mathcal{L}_\text{distill} (\mathbf{p}) = \text{CosineDist} (\mathbf{y}_p, \mathbf{f}_p^{vl}) 
\end{equation}
where $\mathbf{y}_p$ is the distillation prediction by the model for point $\mathbf{p}$. This together with the bounding box regression and the objectness classification losses (based on our auto labels as discussed in Sec.~\ref{sec:det_model_arch}) form our final training objective.

\subsubsection{Open-Vocabulary Inference}
\label{sec:detection_inference}

So far, we have introduced how to train a detector to simultaneously localize all objects in a class-angnostic manner and predict vision-language features for all LiDAR points.
Here, we discuss how we assign open-vocabulary semantics to the predicted boxes during inference. This process is depicted in the right side of Figure~\ref{fig:top_level_overview}. The pre-trained 2D vision language model~\cite{ghiasi2022scaling} contains an image encoder and a text encoder, which are jointly trained to map text and image data to a shared embedding space. As described in Sec.~\ref{sec:distillation}, we add a feature distillation branch that maps 3D input point clouds to the 2D image encoder embedding space, which essentially bridges the gap between point clouds and semantic text queries.
As a result, at the inference time we can encode arbitrary open-vocabulary categories presented as text queries and compute their similarities with the observed 3D points. This can be achieved by computing the cosine similarity between the text query embeddings and the vision-language features predicted by our model for each 3D point. Finally, we assign open-vocabulary categories to boxes based on majority voting. Specifically, we associate each point the category with the highest computed cosine similarity, and then assign to each box the most common category of its enclosing points. 

We would like to emphasize that our approach does not need to process images at inference time, since we have distilled image encoder features to the point cloud. 
Therefore, the only added computation is a simple linear layer for predicting per-point vision-language embeddings, which is negligible compared to the rest of the detector architecture.

\section{Experiments}
\label{sec:exps}

Our UP-VL approach advances the previous state-of-the-art in unsupervised 3D perception for autonomous driving~\cite{najibi2022motion} in two main important directions: 1) enabling open-vocabulary category semantics and 2) detecting objects in all motion states (as opposed to moving-only objects in the previous study). In this section, we perform extensive evaluations with respect to each of these innovations. Note that unsupervised open-set 3D detection is still at early stage in the research community with few published works. Therefore to fairly compare with the state-of-the-art~\cite{najibi2022motion}, we perform our detection experiments first following the same setting as~\cite{najibi2022motion} (\ie detecting class-agnostic moving objects) and then showcasing our new capabilities (\ie detecting objects in any motion states with semantics).

Sec.~\ref{exp:class_agnostic} studies the performance of our system in the class-agnostic setting. This allows us to compare our approach with the existing state-of-the-art method on detecting moving-only objects, showing large improvements. Sec.~\ref{exp:class_aware} moves the needle beyond the capability of the previous class-agnostic state-of-the-art methods and reports results under open-vocabulary class-aware setting for detecting moving-only objects (Sec.~\ref{exp:class_aware_moving}) and the most challenging setting of open-vocabulary detection of objects in all motion states (Sec.~\ref{exp:class_aware_all}). Finally, Sec.~\ref{exp:tracking} reports the open-set tracking quality of our auto labels and Sec.~\ref{exp:qualitative} presents qualitative results. 
See supplementary materials for more ablation studies and error analyses.

\subsection{Experimental Setting}
We evaluate our framework using the challenging Waymo Open Dataset (WOD)~\cite{sun2019scalability}, which provides a large collection of run segments captured by multi-modal sensors in diverse environment conditions. To define moving-only objects in Sec.~\ref{exp:class_agnostic}, we follow~\cite{najibi2022motion} and apply a threshold of 1.0 m/s (\ie $\epsilon^{sf}$ = 1.0). We set the cosine similarity threshold for background categories at $\epsilon^{bg}=0.02$ to achieve best performance in practice. The background categories $C^{bg}$ we exclude from auto labeling are ``vegetation", ``road", ``street", ``sky", ``tree", ``building", ``house", ``skyscaper", ``wall", ``fence", and ``sidewalk".
The WOD~\cite{sun2019scalability} has three common object categories, \ie vehicle, pedestrian, and cyclist. In the class-aware 3D detection experiments (Sec.~\ref{exp:class_aware}), we follow~\cite{najibi2022motion} and combine pedestrian and cyclist into one VRU (vulnerable road users) category, which contains a similar number of labels as the vehicle category. As in~\cite{najibi2022motion}, we also train and evaluate the detectors on a 100m $\times$ 40m rectangular region around the ego vehicle. We use the popular PointPillars detector~\cite{REF:pointpillars_cvpr2018} for all our detection experiments and set an intersection over union, IoU=0.4, for evaluations unless noted otherwise. Please refer to Sec. \textcolor{red}{1} of supplementary materials for a more detailed description of all experimental settings.

\subsection{Class-agnostic Unsupervised 3D Detection of Moving Objects}
\label{exp:class_agnostic}
\begin{table}[!t]
  \centering

    \caption{Comparison of the methods on class-agnostic unsupervised 3D detection of \emph{moving} objects. Top: Auto label boxes. Bottom: Detection boxes.}
      \label{tab:class_agnostic_moving}
      
     \resizebox{\linewidth}{!}{%
    \begin{tabular}{c|c|cc|cc}
    \toprule
     \multirow{2}{*}{Method} & \multirow{2}{*}{Box Type} & \multicolumn{2}{c|}{3D AP@0.4}  & \multicolumn{2}{c}{3D AP@0.5}   \\ 
     & & L1 & L2 & L1 & L2 \\ \midrule

     \eccvname~\cite{najibi2022motion} & \multirow{2}{*}{Auto labels} & 36.9 & 35.5 & 27.4 & 26.4 \\
     UP-VL (ours) &  & \textbf{39.9} & \textbf{38.4} & \textbf{34.2} & \textbf{32.0}   \\
     
     \midrule\midrule
     
     \eccvname~\cite{najibi2022motion} & \multirow{2}{*}{Detections} & 42.1 & 40.4 & 29.6  & 28.4   \\
     UP-VL (ours) &  & \textbf{49.9} & \textbf{48.1} & \textbf{38.4} & \textbf{36.9}   \\
     
     \bottomrule
    \end{tabular}
    }
  \end{table}

For fair comparison, we follow the same setting as~\cite{najibi2022motion} and tailor our approach to class-agnostic moving-only 3D detection. Specifically, we perform auto labeling as introduced in~\ref{method:AML} with speed threshold $\epsilon^{sf}=1.0$m/s and train a class-agnostic detector with feature distillation as described in~\ref{sec:det_model_arch}. However, we disable text queries at inference time. Note that~\cite{najibi2022motion} only considered detection of moving objects. We leave the study of more challenging settings to Sec.~\ref{exp:class_aware}.

\begin{table*}[!ht]
\centering
\caption{Comparison of methods on unsupervised class-aware \emph{moving} object detection. (*since semantics are not available, we report class agnostic AP for the first row, given that vehicle and VRU contain similar number of samples.)}
\label{tab:class_aware_moving}
\begin{tabular}{c|cc|c|cc|c}
\toprule
\multirow{2}{*}{Method} & \multicolumn{2}{c|}{Representations}      &  \multirow{2}{*}{Box type} & \multicolumn{2}{c|}{3D AP} & \multirow{2}{*}{mAP} \\ \cline{2-3} \cline{5-6} 
                        & \multicolumn{1}{l|}{Motion} & Vision-Language &                          & Veh     & VRU    &     \\ \midrule
Clustering~\cite{najibi2022motion}              & \multicolumn{1}{c|}{\checkmark}       &                          & visible                   & N/A         & N/A        & 32.4*       \\
Clustering~\cite{najibi2022motion} + OpenSeg~\cite{ghiasi2022scaling}    & \multicolumn{1}{c|}{\checkmark}       & \checkmark                       & visible                   &  47.8        & 21.5         & 34.7     \\ \midrule
\textbf{Our auto labels}         & \multicolumn{1}{c|}{\checkmark}       & \checkmark     & amodal                          & 57.5        & \textbf{29.8}        & 43.7        \\
\textbf{Our UP-VL detector w. feature distillation}    & \multicolumn{1}{c|}{\checkmark}       & \checkmark                           & amodal                          & \textbf{76.9}         & 28.6        & \textbf{52.8}        \\
\bottomrule
\end{tabular}
\end{table*}

Table~\ref{tab:class_agnostic_moving} shows our result and compares it with \eccvname~\cite{najibi2022motion}. The top part of the table compares the auto labeling quality. The bottom part compares the detector performance between our UP-VL approach and \eccvname. We use the exact same detection backbone and hyper-parameters to ensure a fair comparison. When evaluating at IoU=0.4 as suggested by ~\cite{najibi2022motion}, UP-VL significantly outperforms \eccvname, both in terms of the auto label as well as the detection performance. To better demonstrate our improved auto label quality, we also evaluate with a higher localization criterion at IoU=0.5, where our improvement becomes even more pronounced. We should also point out that in both methods, the final detection quality is superior to the auto label quality. We hypothesize that this is due to the network being able to learn a better objectness scoring function for ranking as well as its ability to denoise the auto labels given the inductive bias of the model~\cite{lowrank21}.

\subsection{Class-aware Unsupervised Open-vocabulary 3D Detection}
\label{exp:class_aware}
In this section, we evaluate the capability of our UP-VL pipeline in class-aware open-vocabulary 3D detection of objects in different motion states. 
Please note that we don't use any 3D human annotations during training and only use the available human labeled categories for evaluation. Moreover, it should be noted that the previous state-of-the-art~\cite{najibi2022motion}, as a class-agnostic approach, falls short in this new setting, making comparisons not possible. In all experiments in this section, we assign labels to boxes by querying category names as text at inference time in an open-vocabulary fashion as described in Sec.~\ref{sec:detection_inference} (see Sec.~{\textcolor{red}{1}} of supplementary for a detailed list of text queries used).

\subsubsection{Moving-only Objects}
\label{exp:class_aware_moving}
Table~\ref{tab:class_aware_moving} reports the class-aware open-vocabulary 3D detection results on the moving-only objects. Since~\cite{najibi2022motion} is no longer applicable in this setting, we construct two baselines for comparison: \ie geometric clustering~\cite{najibi2022motion} which additionally uses our extracted scene flow features ($\mathbf{F}^{sf}_t$) and its variant which leverages both the scene flow features and the vision-language features ($\mathbf{F}^{vl}_t$). 3D point-wise semantics for the baselines are extracted directly by projecting the 2D image features of the pre-trained vision-language model.
We report per-category AP as well as the mAP of these baselines in the top two rows of Table~\ref{tab:class_aware_moving}. The bottom of the table presents the results for our unsupervised auto labels and our final UP-VL detections. Our auto labels and UP-VL detector both outperform baselines constructed from prior approaches. As discussed in Sec.~\ref{sec:distillation}, unlike the baselines that requires applying the image encoder to all camera images at inference time, our detector directly predicts image features extracted by our auto labeling pipeline for 3D point clouds and consequently is more efficient.

\subsubsection{Objects in All Motion States}
\label{exp:class_aware_all}
Finally in this section, we report results on the most challenging setting: unsupervised class-aware open-vocabulary 3D detection for all objects with arbitrary motion states. 
Like Sec.~\ref{exp:class_aware_moving}, since~\cite{najibi2022motion} falls short in this setting, we construct three clustering baselines using different combinations of our features. More specifically, the first row only uses point locations ($\mathbf{P}_t$), the second row uses both point locations and our vision-language features ($\mathbf{F}^{vl}_t$), and the third row leverages all the features including our scene flow features ($\mathbf{F}^{sf}_t$). As an ablation on the effectiveness of the introduced feature distillation in UP-VL, we also add a baseline called ``Our detector w/o feature distillation", where we remove the distillation head and its loss from our detector, and like the baselines in the first three rows, we directly project the vision-language features from camera images to the point cloud for semantic label assignment.
As summarized in Table~\ref{tab:class_aware_all}, our auto labels significantly outperform other baselines listed in the first three rows. 
Moreover, comparing the last two rows, we observe that the proposed vision-language feature distillation leads to significant performance improvement aross all metrics. For example, our approach with feature distillation outperforms the counterpart without distillation by more than 8 points in mAP. 

\begin{table*}[!ht]
\centering
\caption{Comparison of methods on unsupervised class-aware detection of objects in \emph{all motion states}. (*since semantics are not available, we report class-agnostic AP for the first row, given that vehicle and VRU contain similar number of samples.)}
\label{tab:class_aware_all}
\begin{tabular}{c|cc|c|cc|c}
\toprule
\multirow{2}{*}{Method} & \multicolumn{2}{c|}{Representations}      &  \multirow{2}{*}{Box type} & \multicolumn{2}{c|}{3D AP} & \multirow{2}{*}{mAP} \\ \cline{2-3} \cline{5-6} 
                        & \multicolumn{1}{l|}{Motion} & Vision-Language &                          & Veh     & VRU    &     \\ \midrule

Clustering~\cite{najibi2022motion}              & \multicolumn{1}{c|}{}       &                          & visible                   & N/A         & N/A        & 11.6*       \\

Clustering~\cite{najibi2022motion} + OpenSeg~\cite{ghiasi2022scaling}    & \multicolumn{1}{c|}{}       &            \checkmark            & visible                   & 15.8         & 9.9        & 12.9\\

Clustering~\cite{najibi2022motion} + OpenSeg~\cite{ghiasi2022scaling}    & \multicolumn{1}{c|}{\checkmark}       & \checkmark                       & visible                   & 16.1         & 10.0        & 13.1     \\ \midrule

\textbf{Our auto labels}         & \multicolumn{1}{c|}{\checkmark}       & \checkmark     & amodal                          & 30.2         & 14.7        & 22.4        \\

\textbf{Our detector w/o feature distillation}    & \multicolumn{1}{c|}{\checkmark}       & \checkmark                                    & amodal                          & 40.0        & 15.2       & 27.6       \\

\textbf{Our UP-VL detector w. feature distillation}    & \multicolumn{1}{c|}{\checkmark}       & \checkmark                           & amodal                          & \textbf{52.0}         & \textbf{19.7}        & \textbf{35.8}        \\
\bottomrule
\end{tabular}
\end{table*}

\subsection{Tracking}
\label{exp:tracking}
\begin{table}[!ht]
\centering
\caption{Comparison of tracking methods for moving objects with evaluations in class-agnostic (Cls. ag.) and class-aware settings. ``\eccvname-C" refers to class-agnostic {\eccvname} clustering approach, which is unable to be evaluated in the class-aware setting.}
\label{tab:tracking}
\resizebox{\linewidth}{!}{

\begin{tabular}{c|ccc}
\toprule
\multicolumn{1}{c|}{\multirow{2}{*}{Method}} & \multicolumn{3}{c}{MOTA ($\uparrow$) / MOTP ($\downarrow$)} \\ \cline{2-4}
\multicolumn{1}{c|}{}                        & Veh   & VRU   & Cls. ag.    \\ \midrule
\eccvname~\cite{najibi2022motion}            & N/A   & N/A   & 12.8/45.5  \\ \midrule
\eccvname-C~\cite{najibi2022motion}          & \multirow{2}{*}{39.6/37.4}  & \multirow{2}{*}{13.5/53.7}  & \multirow{2}{*}{22.8/43.4} \\ 
+ OpenSeg~\cite{ghiasi2022scaling}           &       &       &            \\ \midrule
UP-VL detector                               & \textbf{65.3}/\textbf{31.0}  & \textbf{24.0}/\textbf{46.8}  & \textbf{41.3}/\textbf{37.4} \\
\bottomrule
\end{tabular}

}
\end{table}

The UP-VL exhibits a high performance not only in detection, but also in tracking - a critical task in autonomous driving. 
We employ the motion-based tracker from~\cite{najibi2022motion}, and conduct experiments in the tracking-by-detection manner. 
We evaluate tracking performance for moving objects and compare our UP-VL detector trained with feature distillation as outlined in Table \ref{tab:class_aware_moving} against two baselines: {\eccvname} detector from Table \ref{tab:class_agnostic_moving} and another open-set baseline from Table \ref{tab:class_aware_moving}. To measure the effectiveness of our model, we employ the widely used MOTA and MOTP metrics, both in the class-agnostic and class-aware open-vocabulary settings. Our experimental results (Table~\ref{tab:tracking}) demonstrate that UP-VL outperforms both baselines by a significant margin.

\subsection{Qualitative Results}
\label{exp:qualitative}
\begin{figure}[!hb]
\vspace{-1em}
\includegraphics[width=\linewidth]{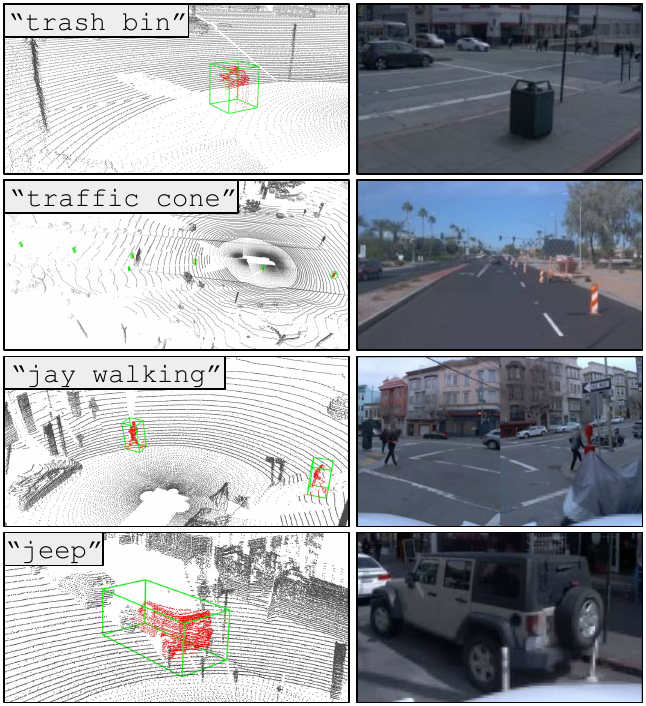}
\caption{Open-vocabulary detection of both static and moving objects via user-provided text queries. Note that in the open-vocabulary setting, the text queries of interested object types are not given in either auto labeling or model training.}

\label{fig:demo}
\end{figure}

Our UP-VL enables open-vocabulary detection of arbitrary object types beyond the few human annotated categories in the autonomous driving datasets. 
Figure~\ref{fig:demo} illustrates some examples. 
In each row, we present the camera image on the right for readers' reference.
On the left, we show the corresponding 3D point cloud and the predicted 3D bounding box by our model based on the open-vocabulary text query provided at inference time.

\section{Conclusions} \label{sec:conclusions}

In this paper, we study the problem of unsupervised 3D object detection and tracking in the context of autonomous driving.
We present a cost-efficient pipeline using multi-sensor information and an off-the-shelf vision-language model pre-trained on image-text pairs.
Core to our approach is a multi-modal auto labeling pipeline, capable of generating class-agnostic amodal box annotations, tracklets, and per-point semantic features extracted from vision-language models.
By combining the semantic information and motion cues observed from the LiDAR point clouds, our auto labeling pipeline can identify and track open-set traffic participants based on the raw sensory inputs.
We have evaluated our auto labels by training a 3D open-vocabulary object detection model on the Waymo Open Dataset without any 3D human annotations.
Strong results have been demonstrated on the task of open-vocabulary 3D detection with categories specified during inference by text queries which we believe opens up new directions towards more scalable software stacks for autonomous driving.

{\small
\bibliographystyle{ieee_fullname}
\bibliography{egbib}
}

\clearpage
\appendix
\section*{Appendix}

\begin{figure*}[!h]
\centering
\setlength{\unitlength}{0.1\linewidth}
\begin{picture}(10, 2.6)
\put(0,0){\includegraphics[width=\linewidth]{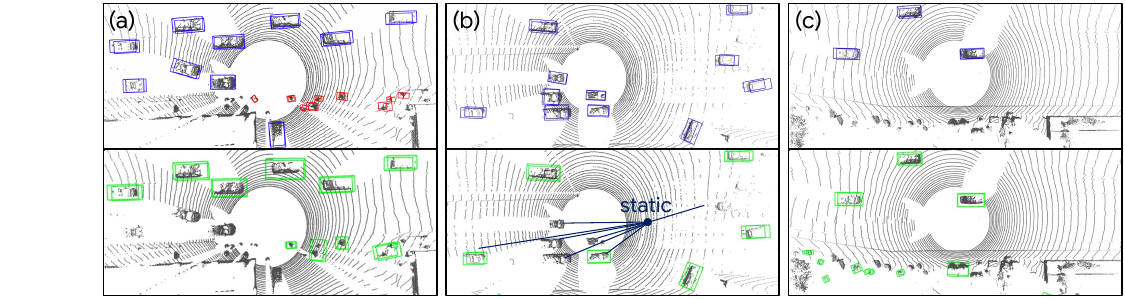}}
\put(0,0.65){MI-UP~\cite{najibi2022motion}}
\put(0,1.95){UP-VL}
\end{picture}
\caption{Comparison of our UP-VL with prior work MI-UP~\cite{najibi2022motion}. Comparatively, our UP-VL (a) localizes objects and classifies them, (b) detects both moving and static objects, (c) produces fewer false positives. Best viewed in color. Box colors: blue for vehicle, red for VRU, green for class-agnostic.}
\label{fig:qual_comparison}
\end{figure*}

\section{Implementation Details}
 This section provides the implementation details for the main two components of our approach namely multi-modal auto labeling, and the open-vocabulary 3D object detector.

\subsection{Unsupervised Multi-modal Auto Labeling}
In our experiments, we use \textsc{vegetation, road, street, sky, tree, building, house, skyscraper, wall, fence} and \textsc{sidewalk} as text queries for defining background categories, $C^{bg}$, which are excluded from auto labeling. We also set the cosine similarities threshold $\epsilon^{bg}$ to be $0.02$. For the experiments in Section~\ref{exp:class_agnostic}, and ~\ref{exp:class_aware_moving} of the main paper which consider moving-only objects, we set a scene flow threshold of $\epsilon^{sf} = 1m/s$ (the same as~\cite{najibi2022motion}).  
For bounding box proposals, we follow Najibi~\etal\cite{najibi2022motion} and set neighborhood threshold to be 1.0m in \texttt{DBSCAN}. 
Without knowing the semantics of objects, it is challenging to define the headings of all objects. 
For moving objects, we align their headings with the object moving direction. For static objects, we choose their headings such that they have an acute angle with the heading of the autonomous driving vehicle.

\subsection{Open-vocabulary 3D Object Detection}
Regarding the vision-language model, in this paper we use the pre-trained OpenSeg model~\cite{ghiasi2022scaling} coupled with the BERT-Large text encoder in Jia~\etal\cite{jia2021scaling} without further fine-tuning on any 2D or 3D autononmous driving datasets.  

For the knowledge distillation, as discussed in Section ~\ref{sec:distillation} of the main paper, we directly distill the final 640 dimensional features of the OpenSeg model.
However, for memory and compute efficiency during training, we first reduce the dimensionality of the features to 64 using an incremental PCA fitted to the whole unsupervised training dataset. 
To evaluate the open-vocabulary detector on the Waymo Open Dataset, we choose the vehicle and VRU as categories of interest, for which the dataset has groundtruth. More specifically, we use \textsc{car, vehicle, parked vehicle, sedan, truck, bus, van, minivan, school bus, pickup truck, ambulance, fire truck} to query for the vehicle category and \textsc{cyclist, human, person, pedestrian, bicycle} to query for the VRU category.
We found that removing queries from this set will lead to dropped mAPs.
For the 3D detection experiments, we use the same two-frame anchor-based PointPillars backbone as previous work~\cite{najibi2022motion} for fair comparisons. We also use the same set of detection losses to train a class-agnostic 3D bounding box regression branch and an objectness score branch, and supplement them with the new distillation introduced in Section ~\ref{sec:distillation} of the main paper. We train models on 64 TPUs, with a batch size of 2 per accelerator. We use a cosine decay learning rate schedule and an initial learning rate of 0.003 and train the models for a total of 43K iterations.

\section{Additional Qualitative Results}
In the paper, we presented qualitative results demonstrating that UP-VL can detect open-set objects using text queries at inference (see Figure~\ref{fig:teaser} and~\ref{fig:demo} of the main paper). Additionally, we included a quantitative comparison with the previous state-of-the-art, MI-UP~\cite{najibi2022motion}, in Table~\ref{tab:class_agnostic_moving} of the main paper. Here, we present qualitative comparison between our UP-VL detector (trained with distillation) and MI-UP~\cite{najibi2022motion} detector in Figure~\ref{fig:qual_comparison}. The top row shows our UP-VL class-aware predictions where the blue and red boxes represent the vehicle and VRU detections respectively. On the bottom, we are showing the class-agnostic predictions of the MI-UP model as green boxes. Comparing column (a), first we can see that unlike MI-UP which is unable to predict semantics, our UP-VL approach can reliably distinguish between objects of vehicle and VRU categories. Moreover, UP-VL can detect many of the objects which were completely missed or grouped together by MI-UP. In column (b), we also mark static objects in the bottom row. Comparing this column highlights another advantage of our approach. While MI-UP is limited to detecting moving-only objects by design, UP-VL is able to detect static objects as well. Lastly, by comparing column (c), one can see that our UP-VL approach can significantly reduce the false positives on cluttered parts of the scene, showing yet another advantage of our approach compared to the prior work on unsupervised 3D object detection in autonomous driving.

\section{Effect of Hyperparameters}
\label{sec:ablation_epsilon}

In this subsection, we perform an ablation study on the effect of the hyper-parameters introduced in Algorithm~\ref{alg:auto_labeling} of the main paper. More specifically, $\epsilon^{bg}$ which is used as a threshold on the computed cosine similarities to define the background points, and $r^{bg}$ which represents a threshold on the required ratio of background points within a box proposal to mark it as background and consequently filtering the proposal. The ablation analysis is presented in Table~\ref{tab:ablation}. First thing to notice is that our approach is fairly robust to these hyper parameters when they are set in a reasonable range. Moreover, comparing the middle rows with the first and last rows demonstrates the effectiveness of introducing these thresholding schemes in improving the mAP of the model. Given these results, in all experiments in the paper we set $\epsilon^{bg} = 0.02$ and $r^{bg} = 0.99$. 

\begin{table}[]
\centering
\caption{Effect of hyperparameters of $\epsilon^{bg}$ and $r^{bg}$.}
\label{tab:ablation}
\resizebox{\linewidth}{!}{
\begin{tabular}{c|cc|c||c|cc|c}
\toprule
\multirow{2}{*}{$\epsilon^{bg}$} & \multicolumn{2}{c|}{3D AP} & \multirow{2}{*}{mAP} & \multirow{2}{*}{$r^{bg}$} & \multicolumn{2}{c|}{3D AP}                         & \multicolumn{1}{c}{\multirow{2}{*}{mAP}} \\ \cline{2-3} \cline{6-7}
                   & Veh          & VRU         &                      &                     & \multicolumn{1}{c}{Veh} & \multicolumn{1}{c|}{VRU} & \multicolumn{1}{c}{}                     \\ \midrule
0.10               & 28.7         & 12.3        & 20.5                 &  50\% & 20.7 &  7.5 & 14.1  \\
0.05               & 29.7         & 14.1        & 21.9                 &  90\% & 27.3 & 11.1 & 19.2  \\
\textbf{0.02}               & \textbf{30.2}         & \textbf{14.7}        & \textbf{22.4}                 &  \textbf{99\%} & \textbf{30.2} & \textbf{14.7} & \textbf{22.4}  \\
0.00               & 29.9         & 14.3        & 22.1                 & 100\% & 30.1 & 14.6 & 22.3  \\ \bottomrule
\end{tabular}
}
\end{table}

\section{Error Analysis}

\begin{figure}[!t]
\includegraphics[width=\linewidth]{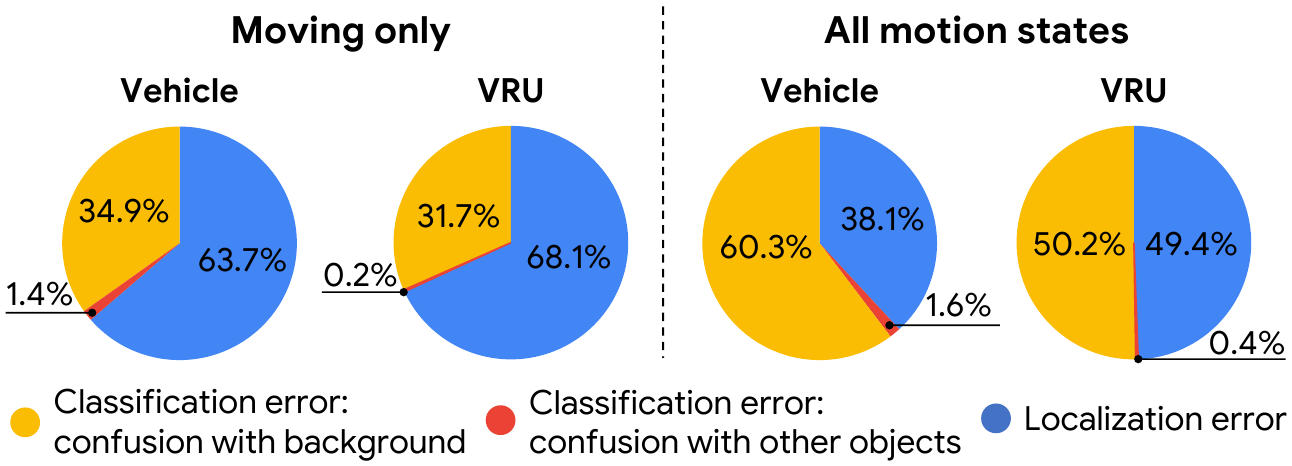}
\caption{Error analysis of false positives. Fractions of false-positives that are caused by classification or localization errors. Our analysis covers two scenarios: detecting moving objects only and detecting objects in all motion states. And we examine both vehicle and VRU categories.}
\label{fig:error_analysis_fp}
\end{figure}

\subsection{Quantitative Analysis}
Section \ref{sec:exps} in the main paper discusses the overall accuracy of our open-vocabulary 3D object detectors. In this subsection, we will delve deeper into the analysis by breaking down the errors. One significant type of errors is false positive detections, which occurs when the detected object does not correspond to any ground truth object, given evaluation thresholds. Following Hoiem \etal ~\cite{hoiem2012diagnosing}, we categorize false positives into three types. \textbf{Localization error} arises when a detected object belongs to the intended category but has a misaligned bounding box (0.1 $<$ 3D IoU $<$ 0.4). The remaining false positives, which have an IoU of at least 0.1 with an ground-truth object from a different category, are classified as \textbf{confusion with other objects}. All other false positives fall under the category of \textbf{confusion with background}. For each category, we count the ``top-ranked" false positives among the most confident $N$ detections, where $N$ is selected to be half the quantity of ground truth objects in that category. Results are presented in  Figure~\ref{fig:error_analysis_fp}. It should be noted that given the decoupled design of our detector, the localization error can be linked to our class-agnostic bounding box prediction branch, and the classification error can be linked to our distillation branch. As can be seen, for moving objects (the left side of the figure), the localization error is the bottleneck in performance. This is while, when we also consider the static objects (the right side of the figure), the share of the classification error noticeably increases. Moreover, as expected, we can see that confusion between the categories (vehicles \vs VRUs) accounts for a very small portion of the false positives. We believe this analysis sheds light on the bottlenecks for further improvements of the proposed approach.

\begin{figure}[!t]
\includegraphics[width=\linewidth]{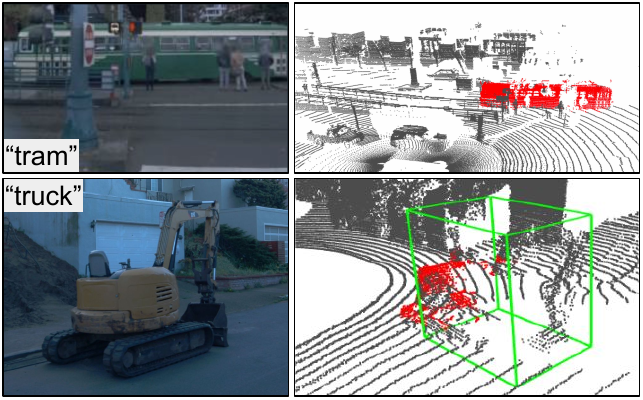}
\caption{Failure cases. (a) Detector fails to generate very large boxes for rare categories like "tram" although the point-wise semantic assignment is correct. (b) Text query of "truck" wrongly matches with an object of crane.}
\label{fig:suppmat_failure}
\end{figure}

\subsection{Qualitative Analysis}
In the previous subsection, we performed quantitative error analysis on the available human annotations in the dataset. Here, we qualitatively present some error patterns of our method in the open-vocabulary setting where human annotations are unavailable. Figure~\ref{fig:suppmat_failure} illustrates some real-world challenges in unsupervised open-vocabulary 3D detection.
One type of failure case is the detector failing to generate a bounding box even though the point-wise cosine similarity has captured the correct semantics from the user's query (\eg ``tram'' in Figure \ref{fig:suppmat_failure}). We believe this is because such kind of large objects are rarely seen in the training data and our detector requires more unsupervised training data to confidently capture those objects. 
Another type of failure case is the mismatch between text queries and visual features for semantically similar concepts. Like the second example in Figure \ref{fig:suppmat_failure}, where a text query of ``truck" has matched with a crane. We hypothesize that this might be due to the similar appearance between cranes and construction trucks and the high co-occurrence of these two object types in the real-world.

\end{document}